\documentclass[conference]{IEEEtran}
\IEEEoverridecommandlockouts
\usepackage{amsmath,amssymb,amsfonts}
\usepackage{algorithmic}
\usepackage{graphicx}
\usepackage{textcomp}
\usepackage{xcolor}

\usepackage{colortbl}
\usepackage{array}
\usepackage{hyperref}  
\definecolor{lightgray}{gray}{0.9}

\begin{document}

\title{Graph Neural Networks for Protein-Protein Interactions - A Short Survey \\
}
\author{Mingda Xu$^{1,2}$, Peisheng Qian$^{1}$, Ziyuan Zhao$^{1}$, Zeng Zeng$^{1}$, Jianguo Chen$^{3}$, Weide Liu$^{4}$ and Xulei Yang$^{1}$
\thanks{This work is supported by Competitive Research Programme "NRF-CRP22-2019-0003", National Research Foundation Singapore. 
$^{1}$Institute for Infocomm Research (I2R), A*STAR, Singapore.
$^{2}$National University of Singapore (NUS), Singapore. 
$^{3}$Sun Yat-sen University (SYSU), China.
$^{4}$Harvard Medical School, United States.
}%
}


\maketitle

\begin{abstract}
Protein-protein interactions (PPIs) play key roles in a broad range of biological processes. Numerous strategies have been proposed for predicting PPIs, and among them, graph-based methods have demonstrated promising outcomes owing to the inherent graph structure of PPI networks. This paper reviews various graph-based methodologies, and discusses their applications in PPI prediction. We classify these approaches into two primary groups based on their model structures. The first category employs Graph Neural Networks (GNN) or Graph Convolutional Networks (GCN), while the second category utilizes Graph Attention Networks (GAT), Graph Auto-Encoders and Graph-BERT. We highlight the distinctive methodologies of each approach in managing the graph-structured data inherent in PPI networks and anticipate future research directions in this domain.
\end{abstract}


\section{Introduction}
The study of interactions between proteins is an essential topic in life sciences. These interactions, serving as essential components in numerous physiological processes within organisms, facilitate our profound understanding of cellular signal transduction, metabolic pathways, and the pathogenesis of various diseases. To predict these interactions more accurately and efficiently, researchers have devoted extensive efforts. While traditional experimental biology methods can provide precise interaction data, they are time-consuming, costly, and require specialized experimental procedures. Moreover, these methods typically focus on the interactions between specific protein pairs, making it challenging to address the demands of large-scale, system-level protein interaction analysis.

With the rapid advancements in artificial intelligence, the potential of machine learning algorithms in predicting protein-protein interactions has progressively gain prominence. Given that protein-protein interaction networks inherently possess graph-structured characteristics, initial predictive methodologies predominantly relied on traditional machine learning algorithms like Random Forests and DeepWalk. However, as the field evolves, deep neural networks have progressively demonstrated their superiority. Especially in recent years, the rise of graph neural networks has further boosted research in this area. Owing to their capability to directly process graph-structured data, graph neural networks are seen as a potent tool for predicting protein-protein interactions. While there have been some surveys~\cite{hu2021survey}~\cite{rasti2019survey} addressing computational methods for protein-protein interaction (PPI) prediction, there remains a gap in the literature regarding a survey that explicitly concentrates on graph-based methods for PPI prediction. This paper aims to review the latest research developments of graph neural networks in forecasting protein-protein interactions, compare the architectures of various methodologies, and explore potential future research directions.

\section{Related Work}

\subsection{PPI Prediction Methods}

Before neural networks became the mainstream in machine learning algorithms, and when hardware capabilities were insufficient to support neural networks, most algorithms used for predicting protein-protein interactions (PPIs) typically relied solely on protein sequence data. Researchers utilized methods such as Substitution Matrix Representation (SMR) \cite{ding2016identification}, Conjoint Triad (CT)~\cite{dey2019classification}, and doc2vec~\cite{yang2020prediction} to extract information from proteins and convert it into uniformly sized sequences or matrices for ease of processing. Building upon this, researchers employed algorithms like Support Vector Machine (SVM) or Random Forests to predict protein-protein interactions.


With the rise in popularity of deep neural networks, an increasing number of studies began to utilize these networks to predict protein-protein interactions, leveraging the enhanced computational power and sophisticated modeling capabilities they offer. Sun \emph{et al.} \cite{sun2017sequence} uses stacked auto-encoder (SAE) to predict PPIs based on protein sequences. A combination of convolutional neural network (CNN) and long short-term memory (LSTM) layers are utilized in the DNN-PPI model \cite{li2018deep}. Yang \emph{et al.} \cite{yang2020prediction-1} uses bidirectional recurrent neural network (Bi-RNN). More recently, graph-based and transformer-based methods~\cite{ijcai2023p554}~\cite{kang2023aftgan} have been proposed for PPI prediction and achieved SOTA performance. 

\subsection{PPI Prediction Datasets}


The STRING database~\cite{string_v_9_1} has undergone multiple updates. By 2019, it had evolved to its $11^{th}$ version, followed by further updates in 2021 and 2023. The database aims to collect and evaluate all publicly available PPI data, refining it through algorithmic estimations. The $11^{th}$ version includes over $5000$ organisms, allowing users to visualize subsets for gene-set enrichment analysis. Beyond the STRING database, researchers frequently utilize its subsets, SHS27k and SHS148k, for various studies.

The Human Protein Reference Database (HPRD)~\cite{keshava2009human} focuses specifically on human protein information. It provides a wide range of data beyond PPIs, such as protein sequences, structures, functions, and their associations with diseases. The Database of Interacting Proteins (DIP)~\cite{DIP} also compiles a diverse array of data, curated by professionals through manual and algorithmic methods to ensure its reliability. 

Moreover, various other protein-protein interaction (PPI) datasets, including HIPPIE~\cite{alanis2016hippie}, Negatome~\cite{blohm2014negatome}, and IntAct~\cite{kerrien2012intact}, are also employed in PPI prediction studies, despite not being as widely used as the previously mentioned datasets.

\section{Graph-based Methods for PPI Prediction}
We classify these approaches into two primary groups based on their model structures. The first category employs Graph Neural Networks (GNN) or Graph Convolutional Networks (GCN), while the second category utilizes Graph Attention Network (GAT), Graph Auto-Encoders (Graph AE) or Graph-BERT. 
\begin{table}[ht]
    \centering
    \caption{Summary of GNN and GCN models}
    \begin{tabular}{>{\columncolor{lightgray}}m{2.1cm} >{\columncolor{lightgray}}m{0.5cm} >{\columncolor{lightgray}}m{1cm} > {\columncolor{lightgray}}m{0.5cm} >{\columncolor{lightgray}}m{0.5cm} > {\columncolor{lightgray}}m{1.8cm}} 
        \hline
        Ref. & Year &  Network & Link & Code & Dataset \\
        \hline
        Liu~\cite{Liu2019IntegratingSA} \cite{Liu2020}  & 2019  & GCN & \href{https://ieeexplore.ieee.org/abstract/document/8983330/}{link} \href{https://bmcbioinformatics.biomedcentral.com/articles/10.1186/s12859-020-03896-6}{link} & N/A & HIPPIE, DIP  \\
        \hline
        Skip GNN~\cite{Huang2020SkipGNNPM}  & 2020  & GNN & \href{https://www.nature.com/articles/s41598-020-77766-9}{link} & \href{https://github.com/kexinhuang12345/SkipGNN}{code} & SHS27k \\ 
        \hline
        Yue~\cite{10.1093/bioinformatics/btz718}  & 2020  & GNN & \href{https://academic.oup.com/bioinformatics/article-abstract/36/4/1241/5581350}{link} & \href{https://github.com/xiangyue9607/BioNEV}{code} & HPRD \\
        \hline
        GNN-PPI~\cite{GNN-PPI}  & 2021  & GNN & \href{https://arxiv.org/abs/2105.06709}{link} & \href{https://github.com/lvguofeng/GNN_PPI}{code} & DIP \\
        \hline
        Kang~\cite{10.1093/bib/bbab513}  & 2021  & GCN & \href{https://academic.oup.com/bib/article-abstract/23/1/bbab513/6456297}{link} & \href{https://github.com/zhanglabNKU/LR-GNN}{code} & STRING \\
        \hline
        Jha~\cite{jha2022prediction} & 2022 & GNN & \href{https://www.nature.com/articles/s41598-022-12201-9}{link} & \href{https://github.com/JhaKanchan15/PPI_GNN.git}{code} & HPRD, DIP \\
        \hline
        HIGH-PPI~\cite{gao2023hierarchical} & 2023 & GNN & \href{https://www.nature.com/articles/s41467-023-36736-1}{link} & \href{https://github.com/zqgao22/HIGH-PPI}{code} & SHS27K, SHS148K, DIP \\
        \hline
        SemiGNN-PPI~\cite{ijcai2023p554} & 2023 & GNN, GCN & \href{https://arxiv.org/abs/2305.08316}{link} & N/A & SHS27K, SHS148K, STRING \\
        \hline
    \end{tabular}
    \label{table1}
\end{table}
\subsection{GNN and GCN for PPI Prediction}
Table~\ref{table1} provides a summary of the research using Graph Neural Networks and Graph Convolutional Networks.

Liu \emph{et al.}~\cite{Liu2019IntegratingSA, Liu2020} were among the early researchers to propose the use of graph convolutional neural networks for predicting protein-protein interactions. They state that traditional prediction methods predominantly rely on amino acid sequence data, neglecting the valuable information embedded within the PPI network. Consequently, this study endeavors to combine PPI network information with amino acid sequence data, leveraging graph convolutional neural networks for prediction. When tested on the HIPPIE and DIP datasets, this approach demonstrated improvements over previous state-of-the-art models. 

SkipGNN model~\cite{Huang2020SkipGNNPM} posits that conventional GNN models are most effective under the assumption that adjacent nodes exhibit direct similarity. However, in the networks of life sciences, the similarity between non-adjacent nodes also holds significant importance. Therefore, the SkipGNN model not only utilizes direct protein-protein interaction information but also leverages indirect second-order interactions. After modifying the original neural network architecture, the authors also employ an iterative fusion scheme to simultaneously use the original neural network and the skip graph. This model demonstrates commendable performance, maintaining robustness even in the presence of noise and data scarcity. 

The work of Yue~\emph{et al.}~\cite{10.1093/bioinformatics/btz718} focuses on the application of graph embedding techniques within biomedical networks. Graph embedding learning is adept at automatically extracting low-dimensional representations of nodes, and traditional methods like matrix factorization have achieved notable success. Motivated by these advancements, the authors aim to apply the latest graph embedding methods to biomedical networks to evaluate their efficacy. Protein-protein interaction networks, as a subset of biomedical networks, have been particularly highlighted. The cutting-edge graph embedding technologies have yielded highly competitive results in this domain.

GNN-PPI ~\cite{GNN-PPI} highlights limited generalizability as a major flaw in current models for predicting protein-protein interactions. This is especially apparent in forecasting interactions involving novel proteins which is often neglected by existing models. To address this, the authors have devised a new evaluation method that places greater emphasis on inter-novel-protein interactions and maintains consistency across different datasets. They introduced the GNN-PPI model, specifically designed to enhance the prediction of novel protein interactions. This model has achieved exceptional results across datasets of various scales, surpassing other models, with its performance being particularly notable in the prediction of inter-novel-protein interactions. 

The LR-GNN model ~\cite{10.1093/bib/bbab513} focuses on representation learning of links to predict inter-molecular relationships. The model harnesses a combination of graph convolutional neural networks and encoders to acquire node embeddings and is equipped with a propagation rule designed to construct link representations from node embeddings. Additionally, the model incorporates a layer-wise fusing rule that combines link representations from all layers in the output, enabling the generation of more accurate results. The protein-protein interaction network is one of the biomedical networks utilized by the authors. Across multiple biomedical network datasets, the LR-GNN model has achieved state-of-the-art results. 

Jha ~\emph{et al.} ~\cite{jha2022prediction} constructs graphs based on PDB files that contain 3D atomic coordinates and extracts node features using protein language models. The authors employed two types of neural network architectures, graph convolutional network and graph attention network, both of which yielded favorable results.


Gao ~\emph{et al.} ~\cite{gao2023hierarchical} contends that existing models are insufficient for capturing the natural hierarchy of protein-protein interactions (PPIs). To address this, this paper introduces a double-viewed hierarchical graph learning model, named HIGH-PPI, for predicting PPIs and inferring molecular details. This model features a hierarchical graph where each node represents a protein graph. In another view, a set of chemically relevant descriptors, as opposed to protein sequences, are utilized as a superior representation of protein relationships. The model is characterized by its high accuracy and robustness.

Zhao~\emph{et al.}~\cite{ijcai2023p554} propose the SemiGNN-PPI model, which employs Mean Teacher and multiple graph consistency regularization to address label scarcity and domain shift in PPI prediction. This model also explores the issue of label dependencies to achieve improved prediction outcomes. Demonstrating exceptional performance on the STRING database, the SemiGNN-PPI model also shows advantages over other models in scenarios with label scarcity.

\begin{table}[ht]
    \centering
    \caption{Summary of GAT, Graph AE, and Graph BERT models}
    \begin{tabular}{>{\columncolor{lightgray}}m{1.9cm} >{\columncolor{lightgray}}m{0.5cm} >{\columncolor{lightgray}}m{1.5cm} > {\columncolor{lightgray}}m{0.5cm} >{\columncolor{lightgray}}m{0.5cm} > {\columncolor{lightgray}}m{1.5cm} } 
        \hline
        Ref. & Year & Network & Link & Code & Dataset \\
        \hline
        S-VGAE~\cite{yang2020graph} & 2020 & Graph AE & \href{https://bmcbioinformatics.biomedcentral.com/articles/10.1186/s12859-020-03646-8}{link} & \href{https://github.com/fangyangbit/S-VGAE}{code} & HPRD, DIP \\
        \hline
        HO-VGAE~\cite{xiao2020graph} & 2020 & Graph AE & \href{https://journals.plos.org/plosone/article?id=10.1371/journal.pone.0238915}{link} & N/A & \href{https://www.nature.com/articles/s41467-019-09177-y}{Kovacs}\\
        \hline
        SN-GGAT~\cite{xiang2021predicting} & 2021 & GAT & \href{https://www.mdpi.com/2218-273X/11/6/799}{link} & N/A & DIP, HIPPIE \\
        \hline
        Struct2Graph~\cite{baranwal2022struct2graph}  & 2022 & GAT & \href{https://link.springer.com/article/10.1186/s12859-022-04910-9}{link} & \href{https://github.com/baranwa2/Struct2Graph}{code} & IntAct, STRING\\
        \hline
        LDMGNN~\cite{zhong2022long}  & 2022  & Transformer, GIN & \href{https://bmcbioinformatics.biomedcentral.com/articles/10.1186/s12859-022-05062-6}{link} & \href{https://github.com/666Coco123/LDMGNN}{code} & SHS27K, SHS148K \\ 
        \hline
        AFTGAN~\cite{kang2023aftgan} & 2023 & Transformer, GAT & \href{https://academic.oup.com/bioinformatics/advance-article/doi/10.1093/bioinformatics/btad052/7000335}{link} & \href{https://github.com/1075793472/AFTGAN}{code} & SHS27K, SHS148K \\
        \hline
        Jha~\cite{jha2023graph} & 2023 & Graph-BERT & \href{https://www.nature.com/articles/s41598-023-31612-w}{link} & \href{https://github.com/JhaKanchan15/PPI_GBERT}{code} & HPRD, \href{https://www.ncbi.nlm.nih.gov/pmc/articles/PMC2808923/}{Negatome}\\
        \hline
        MM-StackEns~\cite{albu2023mm} & 2023 & GAT & \href{https://www.sciencedirect.com/science/article/pii/S0010482522012343}{link} & \href{https://github.com/ alexandraalbu/MM-StackEns}{code} & DIP \\
        \hline
    \end{tabular}
    \label{table2}
\end{table}

\subsection{GAT, Graph AE and Graph BERT for PPI Prediction}

Table \ref{table2} gives a summary of the papers using Graph Attention Networks, Graph Auto-Encoders and Graph-BERT.
Attention mechanisms have revolutionized model designs by eliminating the necessity for recurrence and convolutions. This innovation not only enables the model to selectively focus on specific segments of the input data, facilitating more parallelizable operations, but also significantly reduces training time, leading to superior performance in tasks such as machine translation.

Complementing this, the development of Graph Attention Networks (GATs)~\cite{velickovic2018graph}, offers a robust solution for graph-structured data. GATs adeptly address the limitations of prior graph convolution methods by allowing nodes to attend over their neighborhoods' features, yielding a more dynamic representation. This flexibility in representation, combined with its applicability to both inductive and transductive problems, enables GATs to achieve state-of-the-art results on multiple benchmark datasets.


Baranwal \emph{et al.} \cite{baranwal2022struct2graph} employs graph attention network to forecast protein-protein interactions, enabling the identification of PPIs directly from the structural data of folded protein globules. Struct2Graph, their innovative model, boasts the ability to predict PPI with a remarkable accuracy of 0.9889 on a balanced dataset, comprising an equal number of positive and negative pairs. Impressively, on an unbalanced dataset with a 1:10 ratio between positive and negative pairs, Struct2Graph achieves a fivefold cross-validation average accuracy of 0.9942. Beyond mere prediction, Struct2Graph holds the potential to pinpoint residues that are likely to play a essential role in the formation of the protein-protein complex. The expertise of this residue identification ability was tested across different interaction types, and the model showcased its ability to identify interacting residues with an overall accuracy of 0.87 as well as a good specificity. Similarly, SN-GGAT \cite{xiang2021predicting} and MM-StackEns \cite{albu2023mm} also employ GAT to predict PPIs.

Yang \emph{et al.} \cite{yang2020graph} explores the use of Auto-Encoder for predicting protein-protein interactions. They employ a model named S-VGAE (Signed Variational Graph Auto-Encoder), an enhanced graph representation learning method, to autonomously learn to encode graph structures into low-dimensional embeddings. Experiments on the HPRD and DIP datasets reveal that this model surpasses sequence-based models in terms of accuracy, showcasing it strength in dealing with extremely sparse networks and its ability to generalize effectively on new datasets. Likewise, The HO-VGAE model \cite{xiao2020graph} also utilized graph auto-encoder for PPI predictions.


Zhong \emph{et al.} \cite{zhong2022long} focus their research on long-distance dependency information between two amino acids in sequence and multi-order neighbor information in PPI network. They design a Long-Distance dependency combined Multi-hop Graph Neural Network (LDMGNN) model to solve these two problems. In this paper, they use multi-head self-attention transformer block to encode the amino acid sequences in order to extract the features from them. Also they use two-hop protein-protein interaction (THPPI) network in order to expand the receptive field. They use both PPI and THPPI models as two inputs to an identical graph isomorphic network (GIN) in order to obtain two embeddings. Subsequently they combine the information from both embeddings to predict protein-protein interactions. The model achieved excellent results on both SHS27K and SHS148K datasets.


Kang \emph{et al.} \cite{kang2023aftgan} introduce the Attention Free Transformer and Graph Attention Network (AFTGAN) model, designed for predicting multi-type protein-protein interactions. The ESM-1b embedding utilized in the study has a variety of biological information, including protein sequences, and amino acid co-occurrence similarity. The AFT component is a variant of the transformer architecture that focuses on reducing the complexity and computational demand of traditional attention mechanisms. The GAN component is designed to handle graph-structured data, effectively capturing the relationships and interactions between nodes in a graph. Tested on the SHS27K and SHS148K databases, the AFTGAN model delivered commendable results, outperforming other models in comparison. Notably, it also demonstrated superiority in predicting PPIs between unknown proteins. 

Jha \emph{et al.} \cite{jha2023graph} present a PPI prediction model using a combination of Graph-BERT and the SeqVec language model. The model first uses SeqVec for feature vector extraction from protein sequences, and then employ Graph-BERT for PPI prediction. They employ Graph-BERT because of its effectiveness in tackling prevalent challenges for graph neural networks such as suspended animation and over-smoothing. The model achieves satisfactory results in identifying protein interactions.

\section{Future Work and Conclusion}

Several papers have highlighted the limitations of existing study and suggest various promising avenues for future research.
\cite{Liu2020} suggests there is still space for further optimization of DNN model structures. In~\cite{gao2023hierarchical}, the authors discuss the limitations of hierarchical graph learning for PPI, including its inability to fully utilize protein-level annotations and protein domain information. Additionally, limited memory space restricts the number of views in the hierarchical graph. The model's robustness could also be further improved. Moreover,~\cite{10.1093/bib/bbab513} suggests integrating more graph structural data of biological molecules for better predictions. 

In conclusion, these papers collectively highlight the emerging role of graph neural networks in advancing PPI prediction. The models focus on using graph-based representations to capture complex protein interaction patterns. The models typically employ advanced techniques like attention mechanisms, and integrate diverse biological data to enhance prediction accuracy. However, they share common limitations in dealing with label scarcity, handling domain shifts, and optimizing network architectures for better performance. The future work suggested across these studies involves refining model architectures and enhancing data utilization strategies. These improvements aim to address current challenges and improve prediction accuracy further in the field of PPI prediction using GNN-based approaches.

\bibliographystyle{IEEEbib}
\bibliography{refs.bib}

\end{document}